\documentclass{vgtc}
\graphicspath{{figures/}{pictures/}{images/}{./}}
\usepackage{times}

\usepackage{tabu}
\usepackage{booktabs}
\usepackage{lipsum}
\usepackage{mwe}

\usepackage{mathptmx}

\onlineid{0}

\vgtccategory{Research}

\vgtcinsertpkg

\title{Mind the Gaps: Measuring Visual Artifacts in Dimensionality Reduction}

\author{Jaume Ros\thanks{e-mail: j.ros.alonso@tue.nl} %
\and Alessio Arleo\thanks{e-mail: a.arleo@tue.nl} %
\and Fernando Paulovich\thanks{e-mail: f.paulovich@tue.nl}}
\affiliation{\scriptsize Eindhoven University of Technology}

\abstract{%
    Dimensionality Reduction (DR) techniques are commonly used for the visual exploration and analysis of high-dimensional data due to their ability to project datasets of high-dimensional points onto the 2D plane.
    However, projecting datasets in lower dimensions often entails some distortion, which is not necessarily easy to recognize but can lead users to  misleading conclusions.
    Several Projection Quality Metrics (PQMs) have been developed as tools to quantify the goodness-of-fit of a DR projection; however, they mostly focus on measuring how well the projection captures the global or local structure of the data, without taking into account the visual distortion of the resulting plots, thus often ignoring the presence of outliers or artifacts that can mislead a visual analysis of the projection.
    In this work, we introduce the \textit{Warping Index} (WI), a new metric for measuring the quality of DR projections onto the 2D plane, based on the assumption that the correct preservation of empty regions between points is of crucial importance towards a faithful visual representation of the data. 
}

\usepackage{mathptmx}
\usepackage{amssymb}

\begin{document}

\firstsection{Introduction}

\maketitle

The analysis of high-dimensional data is often hindered by the difficulty of its visualization. A common solution is to use dimensionality reduction (DR) techniques to project the data onto a two-dimensional plane, allowing for direct display as a scatterplot.
DR has become an ubiquitous practice, with many different methods being developed over decades, and whose use has spread far beyond the visualization community.

However, projecting the data into a lower-dimensional space will introduce unavoidable distortions.
There exist a multitude of different techniques, each with distinct hyperparameters that the user needs to fine-tune, with varying strengths and weaknesses, and will produce different results.
Even expert users often struggle to interpret the resulting plots, since it is not trivial to determine which features of the original data have been properly captured in the 2D representation, and how much information has been lost. It is also a possibility that complex structures appear in the projected points, but have no relation to the original data and are merely artifacts of the DR method applied.
A poor understanding of these distortions often leads to incorrect interpretations of the data.

Projection Quality Metrics (PQM) are a common practice in order to quantify how good a projection is, and thus how much its appearance can be trusted. Several alternatives have been developed over the years, focusing on different measures of quality, including the preservation of local neighborhoods and global structures~\cite{Johansson_2009}.

However, these metrics tend to focus on structural aspects of the data, disregarding the visual appearance. This misalignment makes PQMs unable to capture certain types of distortion that, although minor with regards to the structural preservation of the data, produce significant changes in the resulting scatterplot~\cite{Machado_Behrisch_Telea_2025}.
Users are thus given a false sense of security, which may potentially lead to wrong conclusions about the data.

We recognize the need for a PQM that is oriented toward the visual appearance of the projection.
In this work, we present the \textit{Warping Index} (WI), a metric that captures the distortion produced in DR layouts projected in two dimensions. It can be used to visualize the per-region distortion on top of a scatterplot or quantify numerically the projection quality, allowing users to know which visual features of the layout can be trusted.
We complement the explanation with examples of layouts that present particular structures that we know to be artifacts produced by the DR method; however, they cannot be identified through common PQMs.

A Python implementation of the WI is available in the following repository: \url{https://codeberg.org/jros/warping-index}.

\section{Methodology} \label{sec:methodology}

The proposed metric aims to capture the distortion produced in the two-dimensional visual space by comparing the areas of empty regions in the plot to those in the original high-dimensional space~\cite{Aupetit_2007}.

Let $X = x_1, \ldots , x_n \in \mathbb{R}^n$ be a set of high-dimensional points, and $\hat{X} = \alpha(x_1), \ldots , \alpha(x_n) \in \mathbb{R}^2$ their projection into a 2D plane.
We consider a Delaunay triangulation $T(\hat{X})$ of the 2D points that divides the visual space of the projection into a set of triangles $T(\hat{X}) = \hat{t}_1, \ldots , \hat{t}_n$; each triangle $\hat{t}_i$ is composed of three vertices (projected data points) that also define a triangle $t_i$ in the high-dimensional space.
Our proposed metric is based on relating each 2D triangle $\hat{t}_i$ to its high-dimensional counterpart $t_i$ to quantify the distortion of the different regions in the projection space.

The quality of a triangle $Q(\hat{t}_i)$ is thus defined as
\begin{equation}
    Q(\hat{t}_i) = \frac{A(\hat{t}_i)-A(t_i)}{max(A(\hat{t}_i), A(t_i))},
\end{equation}
where $A(t_i) \in (0,1]$ is the normalized area of a triangle $t_i$, once divided by the area of the largest triangle. $Q(\hat{t}_i)$ is in the range $[-1,1]$, where negative values indicate a ``compression'' of the space, positive values indicate a ``stretching'', and a value of 0 means a perfect representation of the area. It can be visualized on top of the scatterplot, as in the examples in Section \ref{sec:results}.

To quantify the distortion of the entire projection $P$, we define the WI as
\begin{equation}
WI(P) = \frac{1}{\sum_{\hat{t}_i \in P} A(\hat{t}_i)} \sum_{\hat{t}_i \in P} A(\hat{t}_i) |Q(\hat{t}_i)|,
\end{equation}
where again a value of 0 indicates a perfect projection, and higher values indicate higher levels of distortion.

Note the weighting of the individual triangle values according to their area in the projected space. This ensures that the metric is bounded in the range $[0,1]$, while giving more importance to large empty regions, which are perceptually more salient.

The computational complexity of the WI is $O(n\ log\ n)$, the cost of the Delaunay triangulation. The amount of triangles to compare is on the order of $O(n)$~\cite{SEIDEL1995115}.
Since the area of a triangle can be computed directly from the lengths of its three sides, our method can also be applied in the absence of explicit embeddings for the high-dimensional points, as long as a distance function obeying the triangular inequality is defined between them (e.g., Euclidean distance).

\section{Results} \label{sec:results}

In this section, we present some relevant cases where the WI proves useful in capturing visual distortion on projected layouts. We consider two distinct datasets and two different DR projections for each, which, although similar in quality according to common PQMs, present a significantly different visual appearance.

\begin{figure}[tbh]
  \centering
  \includegraphics[width=.44\linewidth,trim={2cm 2cm 2cm 2cm},clip]{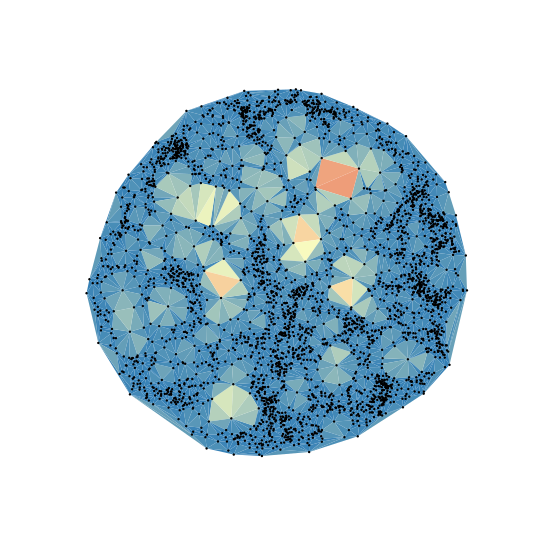}
  \includegraphics[width=.44\linewidth,trim={2cm 2cm 2cm 2cm},clip]{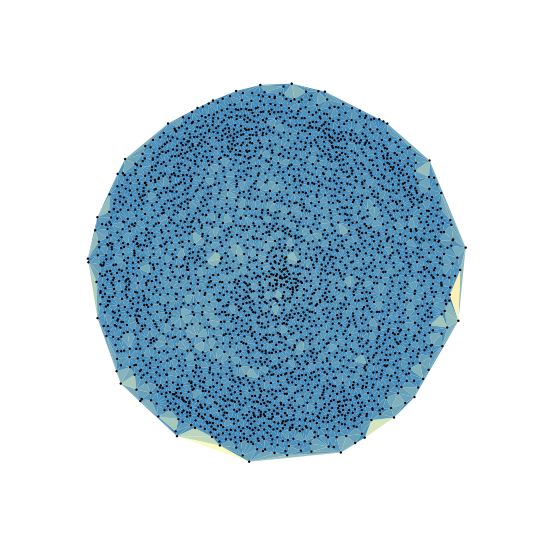}
  \includegraphics[width=0.1\linewidth]{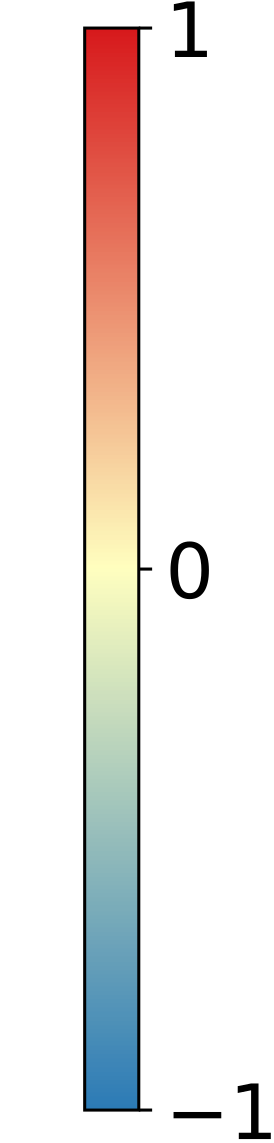}
  \caption{Projection of the \textit{imdb} dataset using FS (left) and GFS (right). The triangulation is colored according to $Q(\hat{t}_i)$. Notice that in both cases there is an overall compression of the space, but it is easy to see that in the left plot some regions have been stretched.}
  \label{fig:imdb}
\end{figure}

The \textit{imdb} dataset has $3,250$ points distributed sparsely in a $700$-dimensional space. Ros et al.~\cite{ros_2025} identified problems when using Force-Scheme (FS) as a dimensionality reduction method due to the appearance of artifacts resembling holes around arbitrary points in an otherwise uniform layout. They introduced Gradient Force-Scheme (GFS) as an improvement to FS that reduced the likelihood of such artifacts, but the usual PQMs could not capture this phenomenon.

Figure \ref{fig:imdb} shows examples of the \textit{imdb} dataset projected with FS and GFS, as presented in Ros et al.~\cite{ros_2025}. The empty regions on the left layout, which were found to be an artifact produced by FS, can easily be identified as such when visualizing the quality of the triangles $Q(\hat{t}_i)$. On the right layout, it can be seen that, in fact, all the areas have been compressed with respect to the original space, indicating that the high-dimensional space cannot properly fit in two dimensions.

\begin{table}[th]
\caption{Quality of the projections of the \textit{imdb} obtained with FS and GFS, measured by stress, trustworthiness, and the WI.}
\label{tab:imdb}
\centering
\begin{tabular}{llll}
\textbf{Method} & \textbf{Stress} ($\downarrow$) & \textbf{Trust.} ($\uparrow$) & \textbf{Warping Index} ($\downarrow$) \\ \hline
FS        & 0.2564 & 0.5663          & 0.9574                     \\
GFS       & 0.2713 & 0.5681          & 0.8966                    
\end{tabular}
\end{table}

Just by observing the quality of the triangles, it is hard to determine which of the two layouts is better in this case. However, by computing the WI, it can be seen that GFS indeed produces a less perceptually distorted plot, as shown in Table \ref{tab:imdb}. Notice how stress and trustworthiness are not able to capture the distortion in the left layout; in the case of stress, the left layout would even be preferred.

\begin{figure}[tbh]
  \centering
  \includegraphics[width=.44\linewidth,trim={2cm 2cm 2cm 2cm},clip]{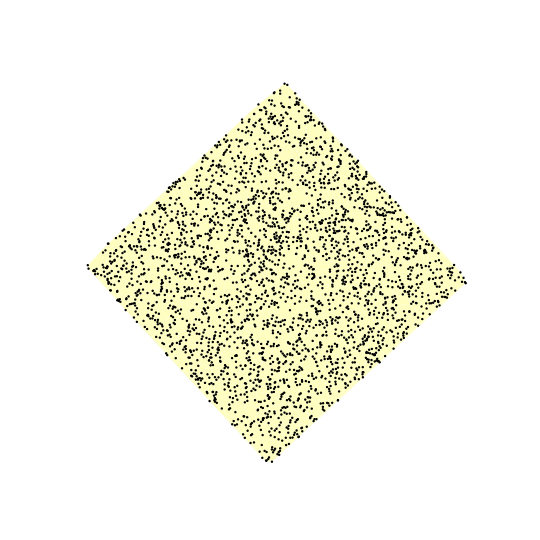}
  \includegraphics[width=.44\linewidth,trim={2cm 2cm 2cm 2cm},clip]{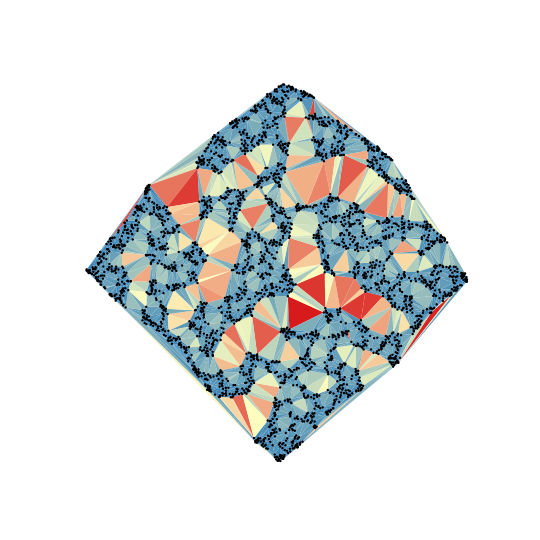}
  \includegraphics[width=0.1\linewidth]{figs/color_bar.png}
  \caption{Projection of the \textit{square} dataset. PCA (left) shows minimal distortion (most of the triangles have a $Q(\hat{t}_i)$ value of 0), while t-SNE (right) can show non-existing empty regions, captured by $Q(\hat{t}_i)$.}
  \label{fig:square}
\end{figure}

The \textit{square} dataset is a synthetic one that showcases the downside of local DR methods such as t-SNE. It consists of three thousand randomly distributed 3D points in the range $[0,1]\times[0,1]\times[0,0.001]$ (i.e., in a 2D plane with some noise in the third dimension), which should be easy to plot in 2D by just ignoring the third component. PCA does precisely that, but t-SNE (with a default perplexity value of $30$) tends to create clusters and give a false visual impression of empty regions in the data, as can be seen in Figure \ref{fig:square}.

However, once again, stress and trustworthiness are barely able to capture this distortion, as shown in Table \ref{tab:square}. Only the WI significantly captures the distortion introduced by t-SNE's clustering.

\begin{table}[th]
\caption{Quality of the projections of the \textit{square} dataset obtained with PCA and t-SNE, measured by stress, trustworthiness, and the WI.}
\label{tab:square}
\centering
\begin{tabular}{llll}
\textbf{Method} & \textbf{Stress} ($\downarrow$) & \textbf{Trust.} ($\uparrow$) & \textbf{Warping Index} ($\downarrow$) \\ \hline
PCA        & 0.0000 & 1               & 0.0010                     \\
t-SNE      & 0.0196 & 0.9999          & 0.7758                    
\end{tabular}
\end{table}

In the previous examples, it can be seen how common PQMs, which rely on measuring the preservation of high-dimensional structure, can be inadequate in capturing some artifacts that are, however, important for a visual analysis of the projection.

\section{Conclusion \& future work}

In this poster, we examined the limitations of conventional PQMs in capturing the visual distortion present in 2D scatterplots of dimensionally-reduced data.
We propose the WI as a PQM that can capture such distortions by focusing on the areas of the projected space rather than individual points and their relationships.

Future work can focus on improving the definition of the WI by exploring other triangulation methods or even considering if it is possible to segment the space into regions other than triangles. It is also interesting to consider other types of distortion beyond compression/expansion of the space.

As mentioned in Section \ref{sec:methodology}, the WI does not require explicit embeddings in the high-dimensional space: a distance metric between points is sufficient. However, the requirement of a distance that satisfies the triangular inequality can limit its application in some cases. Adapting the WI to work without this restriction is also an interesting direction for future research.

\bibliographystyle{abbrv-doi-hyperref}

\bibliography{bibliography}

\end{document}